\documentclass[11pt,letterpaper]{berkeley}

\usepackage[numbers,compress]{natbib}

\usepackage[utf8]{inputenc} % allow utf-8 input
\usepackage[T1]{fontenc}    % use 8-bit T1 fonts
\usepackage{hyperref}       % hyperlinks
\usepackage{url}            % simple URL typesetting
\usepackage{booktabs}       % professional-quality tables
\usepackage{amsfonts}       % blackboard math symbols
\usepackage{nicefrac}       % compact symbols for 1/2, etc.
\usepackage{microtype}      % microtypography
\usepackage{xcolor}         % colors
\usepackage{amsmath}
\usepackage{amssymb}
\usepackage{mathtools}
\usepackage{amsthm}
\usepackage{enumitem}

\usepackage{arydshln}
\usepackage{pifont}

\usepackage{color-edits}
\addauthor{vc}{blue}
\addauthor{vince}{red}
\addauthor{sw}{purple}

\usepackage[capitalize,noabbrev]{cleveref}

\theoremstyle{plain}

\theoremstyle{definition}

\theoremstyle{remark}

\title{Align AI to Dynamic Human-AI Workflows}

\author{
Valerie Chen$^{1}$, Cleotilde Gonzalez$^{1}$, Anita Williams Woolley$^{1}$,
Michael Lee$^{2}$, Tongshuang Wu$^{1}$, Vincent Conitzer$^{1}$,
Aarti Singh$^{1}$\\
$^{1}$Carnegie Mellon University\qquad
$^{2}$University of California, Irvine
}

\begin{abstract}
Current alignment approaches typically focus on emulating human behavior using static representations of human preferences, failing to capture the dynamic, context-dependent nature of real-world human-AI interactions. 
In this paper, we argue for a shift from static and emulative to interactive and complementary alignment, where preferences emerge through interaction and alignment is defined not by satisfying preferences alone.
We first formalize this gap by contrasting existing alignment with a trajectory-level view in which human and model behavior co-evolve over time. 
Because these interaction dynamics have not been adequately captured within existing ML formulations, we ground this perspective in insights from an interdisciplinary workshop.
We draw on lessons from social-science accounts of human-human collaboration and then argue that human-AI systems amplify these dynamics, introducing new asymmetries that make reasoning about uncertainty harder and introduce new coordination challenges.
Based on these lessons and new challenges, we conclude by outlining a research agenda for developing AI systems that align with humans in interaction, requiring an interdisciplinary synthesis of machine learning and the social and decision sciences.
\end{abstract}

\begin{document}

\maketitle

\section{Introduction}

\textit{Consider a developer using an AI coding assistant. The developer initially relies on suggestions provided by the AI assistant to move quickly, but after encountering incorrect outputs, the developer becomes more cautious, preferring explanations and verifying results more carefully. As the assistant proves reliable again, their reliance increases and supervision might reduce.}

This example of human-AI collaboration reflects not only changing preferences but also evolving trust and reliance shaped by a user's interaction history with an AI system. 
It also highlights how the AI assistant serves not just to emulate human-like behavior; rather, humans and AI should coordinate, adapting roles to leverage complementary strengths \citep{Gonzalez2025-ob}. 
Building AI systems that can work effectively with people, therefore, requires accounting for these dynamics.
Despite these interaction-driven dynamics, most alignment techniques for AI models focus on emulating or mimicking humans using static preferences or demonstrations (e.g., as in preference-based RL~\citep{wirth2017survey,xu2020preference}, dueling bandits~\citep{yue2009dueling}, behavior cloning~\citep{pomerleau1989alvinn,argall2009survey,ross2011dagger}).
This paradigm persists in modern alignment methods, which primarily target conversational behavior and verbal outputs through preference optimization~\citep{christiano2017deep,ziegler2019fine,Stiennon2020-hf,rafailov2023direct}.
For example, reinforcement learning from human feedback (RLHF) optimizes models using stand-alone comparisons~\citep{ouyang2022training}. 
% These methods may be appropriate for when they are being used as chatbots, as they frequently are.  

\begin{figure}[t]
\centering
\includegraphics[width=\textwidth]{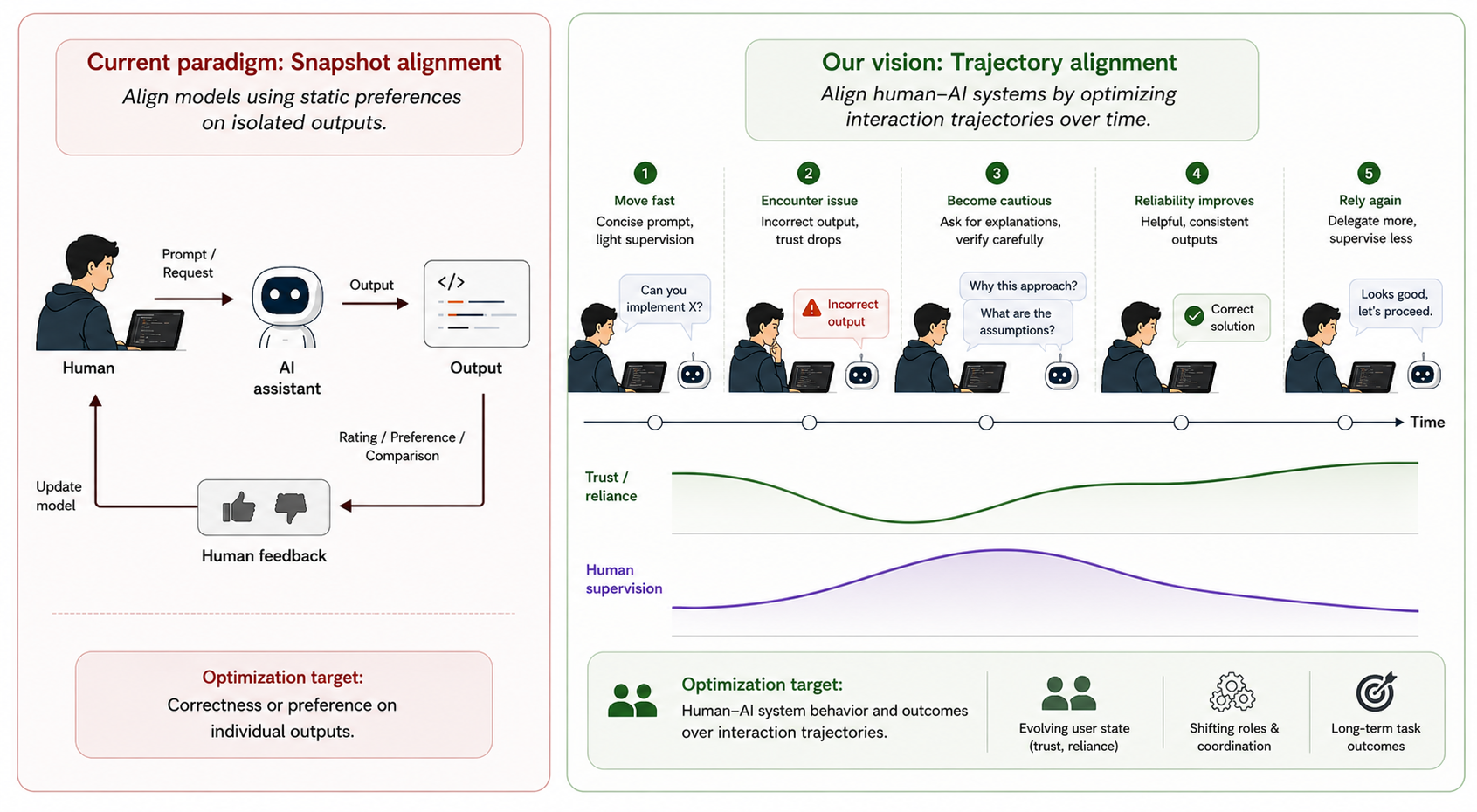}
\caption{\textbf{Rethinking AI Alignment for Human-AI Workflows.} Current alignment paradigms optimize models using static preferences to emulate human outputs. This paper argues for shifting from snapshot evaluation and optimization toward a dynamic, complementary view of alignment, where the goal is to optimize the behavior of human-AI systems over sequences of interaction.}
\label{fig:motivation}
\end{figure}

As agentic AI systems are increasingly taking actions in the world and integrating into human workflows~\citep{Jarrahi2018-cd,Maedche2019-se,Kellogg2020-pr,Shneiderman2020-xw,Horvitz1999-mi,Parasuraman2000-levels}, this shift highlights a fundamental gap in how alignment is currently defined and evaluated.
It is not evident that the kinds of static and emulative alignment for conversational behavior and verbal outputs will prove effective or appropriate for more agentic systems~\citep{shen2026scaling}.  
In this paper, we argue that this gap requires moving towards methods that capture longitudinal interactions (Figure~\ref{fig:motivation}). 
Doing so raises a set of open challenges:\textit{ what does it mean for AI systems to complement human collaborators and how can such behaviors be optimized in practice?}

Because these challenges are not well understood solely from a machine learning perspective, we draw on insights synthesized from a cross-disciplinary workshop spanning machine learning and the social sciences.
First, we draw on prior social science literature on human-human collaboration as a foundation for understanding how trust and complementarity emerge in multi-agent interaction.
We then discuss how human-AI collaboration extends these foundations but under new conditions: AI systems amplify asymmetries in capability and information in ways that are often difficult for users to interpret, while lacking the social and institutional cues that support trust and accountability in human teams. AI systems can also take on new roles that reshape how joint work is structured.
Finally, we identify new challenges for alignment in interaction that are unique to human-AI collaboration.

\textbf{In this paper, we argue for moving beyond optimizing snapshot model performance to optimizing trajectories of human-AI systems over time, drawing on interdisciplinary perspectives to better characterize the dynamics of human-AI interaction.}
While we introduce foundational considerations for aligning AI systems in human workflows, many open questions remain. To chart concrete paths forward, we identify key conceptual, translational, and evaluation challenges, and conclude with takeaways for the human-AI alignment community.

\section{Background}

% Requires:
% \usepackage{arydshln}

\begin{table}[t]
\centering
\caption{\textbf{Standard alignment methods optimize static objectives and omit key properties of collaborative interaction.}
Here, $x$ denotes an input, $y$ a model output, $\pi_\theta$ a learned policy, and $r_\phi$ a learned reward model. We contrast these methods with a collaborative objective over trajectories $\tau$, where reward depends on interaction history $\tau_{<t}$ and joint human-AI outcomes.}

\resizebox{\linewidth}{!}{
\begin{tabular}{p{0.5\linewidth} p{0.4\linewidth}}
\toprule
\textbf{Method} & \textbf{Objective} \\
\midrule

Behavior cloning~\citep{pomerleau1989alvinn,argall2009survey,ross2011dagger}
& $\max_{\pi_\theta} \log \pi_\theta(y^H \mid x)$ \\

\cdashline{1-2}[0.4pt/1.5pt]
\multicolumn{2}{p{0.90\linewidth}}{\emph{Limitation:} $\pi_\theta$ learns to imitate human outputs, without capturing how interaction history shapes collaborative outcomes over time.} \\

\midrule

Preference-based alignment~\citep{wirth2017survey,ziegler2019fine,ouyang2022training,rafailov2023direct} & $\max_{\pi_\theta}
\mathbb{E}_{y \sim \pi_\theta(\cdot \mid x)}
[r_\phi(x,y)]$\\

\cdashline{1-2}[0.4pt/1.5pt]
\multicolumn{2}{p{0.90\linewidth}}{\emph{Limitation:} $\pi_\theta$ may only learn to reflect short-term user satisfaction rather than complementary behavior throughout longer interaction trajectories.} \\

\midrule

Dueling bandits~\citep{yue2009dueling}
& $\max_{\pi_\theta}
\mathbb{E}_{(y,y') \sim \pi_\theta}
\left[\mathbf{1}(y \succ y')\right]$\\
\cdashline{1-2}[0.4pt/1.5pt]
\multicolumn{2}{p{0.90\linewidth}}{\emph{Limitation:} Optimizes for outputs that win pairwise comparisons, while abstracting away coordination, shared context, and downstream collaborative effects.}\\

\midrule

\textbf{Collaborative alignment}
& $\max_{\pi_\theta} \mathbb{E}_{\tau}\left[\sum_t r^*(x_t, a_t^H,a_t^{AI};\tau_{<t})\right]$ \\

\cdashline{1-2}[0.4pt/1.5pt]
\multicolumn{2}{p{0.90\linewidth}}{\emph{Goal:} Optimize human-AI interaction trajectories directly, where rewards depend on interaction history, coordination, complementary expertise, and joint outcomes.} \\

\bottomrule
\end{tabular}
}
\label{tab:alignment-methods-user-state}
\end{table}

\subsection{The Limitations of Current Alignment Approaches}

We first review the dominant paradigm underlying many modern alignment approaches. 
Let $x \in \mathcal{X}$ denote an input, such as a prompt or context, and let $y \in \mathcal{Y}$ denote a model output. 
Alignment is often operationalized through a preference function $r^*(x,y)$ that assigns a scalar value reflecting human judgment of the output.

In practice, $r^*$ is not directly observed, but is approximated using human feedback, such as ratings or pairwise comparisons over outputs. 
This yields a learned reward model $\hat{r}(x,y)$, which is then used to optimize a policy $\pi_\theta(y \mid x)$:
\begin{equation}
\max_{\pi_\theta} \ \mathbb{E}_{x \sim \mathcal{D}, \, y \sim \pi_\theta(\cdot \mid x)} \left[ \hat{r}(x, y) \right].
\end{equation}

This formulation has been productive for improving individual model outputs, but it embeds several assumptions that become limiting in real-world human-AI systems (see Table~\ref{tab:alignment-methods-user-state}).
First, preferences are treated as static, without incorporating interaction history.
Second, alignment is defined at the level of individual outputs rather than as a collaborative process.
Third, the objective is specified in terms of a reward over model outputs alone, rather than a joint reward over human-AI interaction.
As a result, the human is primarily represented as a source of preference labels, while the AI system is optimized to emulate or satisfy an inferred preference function.

\subsection{Formalizing Alignment for Collaborative Settings}

To address the limitations of current approaches, we move from isolated inputs and outputs to modeling sequential interaction between agents. 
Let the interaction trajectory be
\[
\tau = (x_1, a_1^A, a_1^B, x_2, a_2^A, a_2^B, \dots, x_T,a_T^A, a_T^B),
\]
where $x_t$ denotes the observable interaction context at time $t$, and $a_t^A, a_t^B$ are the actions taken by two collaborators. 
At each step, actions are conditioned on the preceding interaction history $\tau_{<t}$ as well as latent collaborator state $z_t$, such as goals, beliefs, expertise, or mental models that may not be directly observable~\citep{Rabinowitz2018-tom,Carroll2019-human}. 
Rather than optimizing for a fixed input $x$, this formulation models collaboration as an evolving process in which behavior depends on both observable interaction context and unobserved internal state over time.

In this setting, reward is also defined as a function of how the interaction unfolds over time:
\[
r_t = r^*(x_t, a_t^A, a_t^B; \tau_{<t}).
\]

We can also capture \emph{complementarity} between collaborators, which arises when
\[
r^*(x_t,a_t^A, a_t^B; \tau_{<t}) > \max \big( r^*(x_t,a_t^A, \emptyset;  \tau_{<t}), \; r^*(x_t,\emptyset, a_t^B; \tau_{<t}) \big). 
\]
This formulation captures both human-human and human-AI collaboration.\footnote{We note that prior work on human-AI collaboration only considers the one-time-step notion of complementarity~\citep{bansal2021updates,rastogi2023taxonomy,Vaccaro2024-te}.} In human-human settings, both agents adapt to each other through interaction, but their behaviors are not directly controllable or optimized by a system designer.
In human-AI settings, however, one of the agents is a parameterized system with policy $\pi_\theta$, where:
\[
    a_t^{AI} \sim \pi_\theta(\cdot \mid \tau_{<t}),
\]
and the policy $\pi_\theta$ can be optimized.

In human-AI collaboration, the AI system is a parameterized system with policy $\pi_\theta$ that can be optimized:
\[
\max_{\pi_\theta} \ \mathbb{E}_{\tau \sim \pi_\theta} \left[ \sum_{t=1}^T r^*(x_t,a_t^H, a_t^{AI}; \tau_{<t}) \right],
\]
where the trajectory distribution depends jointly on human and AI actions. 
We concretize our discussion using a well-known example: consider a developer working with an AI coding assistant. 
The human actions $a_t^H$ may include prompting, verifying outputs, or editing generated code, while AI actions $a_t^{AI}$ may include generating code or expressing uncertainty before proceeding.

We observe that optimizing $\pi_\theta$ is not simply a matter of maximizing immediate user preference because an action that is preferred in the moment may still produce poor downstream trajectories, for example if it increases overreliance on the AI system~\citep{parasuraman1997humans,lee2004trust,bjork2011making}. 
Conversely, actions such as expressing uncertainty or requesting clarification may improve the long-run human-AI workflow even if they are locally less preferred. 
For AI-assisted coding, the relevant outcome is therefore not only whether users accept generated code, but whether the joint workflow produces correct, maintainable code while preserving appropriate human oversight~\citep{mozannar2024realhumaneval}.

\section{Interdisciplinary Perspectives on Human-AI Alignment}

To make progress on building AI systems that can take actions that complement users in the long term, we argue that this research agenda cannot be addressed by the machine learning community alone.
It requires integrating complementary perspectives from fields such as human-computer interaction, cognitive science, and organizational psychology, which have long studied how humans collaborate over time and are beginning to shed light on the nuances of human-AI collaboration~\citep{suchman1987plans,hollan2000distributed,amershi2019guidelines,gonzalez2026toward}.
We therefore convened an interdisciplinary workshop to answer these questions. 

\subsection{Methodology}

The theme of the workshop convening was creating flexible Human-AI teams to achieve complementarity, which called for participants interested in the interdisciplinary study of how to design and deploy AI systems in ways that are dynamically aligned with human values, robust to unexpected behavior, and safe even under failure modes.
We now discuss the details of the workshop held in September 2025:

\textbf{Participants.}
Participation was determined through an application process intended to select a diverse group of attendees in terms of disciplinary background, methodological expertise, career stage, and application domain. 
In total, 70 participants attended the workshop, representing 43 institutions across 8 countries. 
Participants included professors, postdoctoral researchers, industry research scientists, and graduate students. 
Their backgrounds were roughly split evenly between machine learning and social sciences.

\textbf{Pre-workshop survey.}
Prior to the workshop, all participants completed a survey used to identify discussion topics for in-workshop activities. 
The survey focused on two dimensions: future directions and barriers to progress.

\textbf{Workshop procedure.}
The workshop took place over two days. 
The first half focused on tutorials, talks, and posters, establishing shared context across disciplines. 
The second half focused on structured brainstorming and discussion. 
Each topic was led by two experts with complementary expertise spanning machine learning and social or decision sciences. 
Participants were divided into six groups with roughly balanced disciplinary representation, and groups rotated across topics to ensure broad participation.

\begin{figure}[t]
\centering
\includegraphics[width=\textwidth]{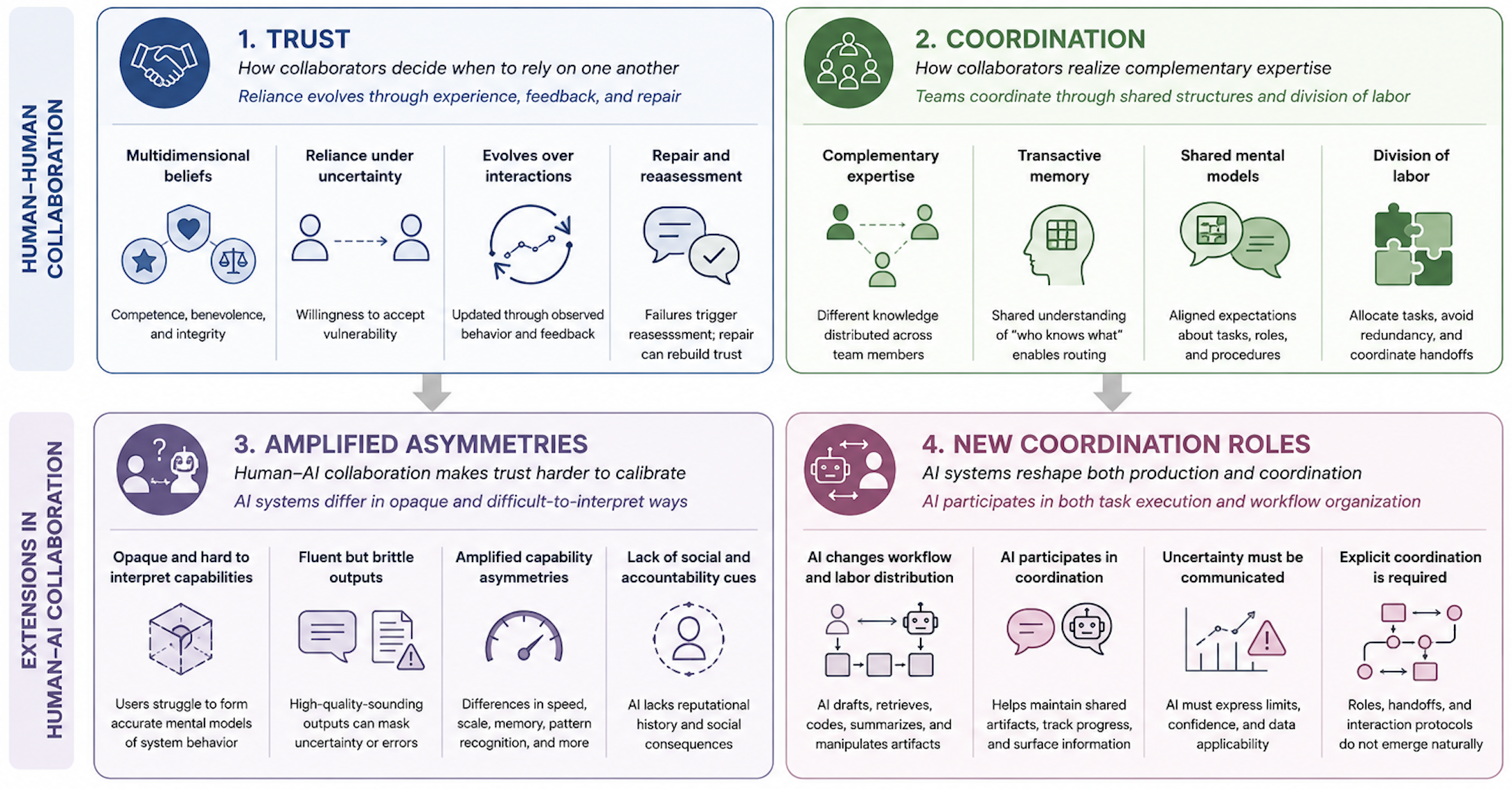}
\caption{\textbf{From Human-Human Collaboration to Human-AI Alignment.} We draw on literature studying human teamwork and collaboration and identify ways in which existing principles change in human-AI settings.}
\label{fig:findings}
\end{figure}

\textbf{Presentation of findings.} The findings presented in this paper are synthesized from the second day of the workshop, including topic discussions, structured rotations, in-depth brainstorming sessions, and written summaries from facilitators and participants.
Across these activities, several recurring themes emerged. Participants repeatedly pointed to insights from human-human collaboration as a foundation for understanding effective interaction, which we draw on to identify key principles.
They also highlighted ways in which these principles change in human-AI settings.
Our findings are summarized in Figure~\ref{fig:findings}.

\subsection{Lessons from Human-Human Interaction}\label{sec:human-human}

We use human-human collaboration as the starting point because social science has long studied how humans coordinate with one another. 
The workshop discussion repeatedly returned to two bodies of work that are especially relevant for dynamic and complementary alignment: research on trust as an evolving relational state, and research on team cognition as a mechanism for coordinating complementary expertise~\citep{Kozlowski2006-wz,Marks2001-zc}.

\textbf{Lesson 1: A user's trust in their collaborator is multidimensional and changes over interactions.}
In collaborative settings, individuals must decide whether and how much to rely on a partner, despite not fully knowing their partner's full competence, intentions, or future behavior. 
Trust is commonly referred to as a willingness to accept vulnerability under these types of uncertainty~\citep{Rousseau1998-tk}.
Prior literature shows that trust is multidimensional, comprising beliefs about a collaborator’s competence, benevolence, and integrity~\citep{mayer1995integrative,schoorman2007integrative,legood2023critical}. 
Perceived competence primarily determines whether individuals delegate tasks or accept a partner’s outputs, while perceived benevolence and integrity shape whether they are willing to share information in the collaborative process~\citep{Dirks2001-jm}. 
Behavioral experiments show individuals will typically continue to rely on partners who are capable but occasionally make mistakes, but will rapidly withdraw when they infer deception or misaligned intent~\citep{mcallister2006trust,Lewicki2006-ip,dietvorst2015algorithm,Logg2019-appreciation}.

Trust also evolves through interactions and is updated based on observed behavior, feedback, and repair attempts~\citep{mayer1995integrative,yu2014developing,Lewicki2006-ip}. 
Computational and cognitive models characterize trust as emerging through repeated interaction, reinforcement, expectation formation, and reciprocal adaptation~\citep{gonzalez2015cognitive,juvina2013reciprocal,harman2014dynamics}. 
Longitudinal studies of team interactions show that early impressions quickly give way to experience, with consistent performance strengthening reliance and failures triggering reassessment~\citep{mcallister2006trust}.

\textbf{Lesson 2: Collaboration depends on complementary expertise, but realizing these gains requires active coordination.}
While trust governs when and how individuals rely on a collaborator over time, a separate challenge is identifying how collaborators can effectively coordinate and divide labor to leverage complementary expertise.
This process requires coordination~\citep{Steiner1972-tw,Hackman2002-he,Malone1994-bp,Crowston1997-bb} and, in practice, groups often fail to do this well. 
For example, classic studies of group decision-making show that teams tend to focus on information that is already shared among members, while neglecting unique, and potentially critical, information held by individuals~\citep{Stasser1985-ok,Wittenbaum1996-ch}. 
Related work shows that coordination may lead to production blocking (i.e., because group members cannot contribute simultaneously, some ideas are delayed, forgotten, or never voiced)~\citep{Diehl1987-hj,Mullen1991-fx}, evaluation apprehension (i.e., one's hesitation to share ideas that may be judged)~\citep{Diehl1987-hj,Mullen1991-fx}, and premature consensus (i.e., early convergence without full exploration)~\citep{Janis1982-kl}. 

Effective teams overcome these challenges by developing shared structures that guide reliance. One key mechanism is \emph{transactive memory}, or a shared understanding of “who knows what” within the group~\citep{Wegner1987-nz,Lewis2003-il}. This allows members to route questions, divide labor, and access expertise without redundant effort, enabling more efficient coordination~\citep{Liang1995-re,Ren2011-nw,Argote2012-vl}. For example, in software teams, developers often specialize in different components of a system, and effective collaboration depends on knowing which teammate to consult or defer to for a given issue~\citep{Faraj2000-cb,Herbsleb2003-vl}.
A complementary mechanism is the development of \emph{shared mental models}, which align members’ expectations about tasks, roles, and procedures~\citep{Klimoski1994-xi,Mathieu2000-ye,Cannon-Bowers2001-iu}. These shared expectations reduce the need for constant communication and enable smoother coordination, particularly in dynamic or high-stakes settings~\citep{Marks2001-zc,Edmondson1999-bt,Woolley2010-qy}. In aviation and military teams, shared mental models help crew members anticipate handoffs, monitor threats, and adapt under time pressure~\citep{Salas1995-ok,Salas2008-gb}; in medical trauma teams, shared expectations about roles and protocols help specialists coordinate rapidly while still adapting when the case violates routine assumptions~\citep{Faraj2006-mg}.

\subsection{Revisiting Lessons in Human-AI Collaboration}

We revisit lessons from human-human collaboration to examine how these principles are reshaped in the context of human-AI collaboration.

\textbf{Extension 1: Trust must be formed under amplified and less interpretable asymmetries.}
Human-AI workflows amplify the challenge of reasoning about uncertainty in collaboration because asymmetries between collaborators are larger and qualitatively different. 
AI systems may differ from humans in speed, scale, memory, statistical pattern recognition, confidence calibration, and vulnerability to distribution shift~\citep{Shneiderman2020-xw,Dafoe2021-ea,Bainbridge1983-ironies,Parasuraman2000-levels}. 
While users often recognize that AI systems behave differently from human collaborators, prior work shows that they struggle to form accurate mental models of system behavior and to calibrate reliance appropriately~\citep{Norman1990-automation}. 
For example, explanations and decision aids are intended to make model behavior more transparent, yet they do not reliably help users determine when a system will succeed or fail, and can lead to over-reliance or under-utilization~\citep{Bansal2019-zt,buccinca2021trust,Miller2019-vi,10.1145/3610219}. 

More broadly, the absence of familiar social and institutional cues makes AI systems harder to interpret as collaborators. 
For example, studies of human-AI interaction show that users often form incomplete or incorrect mental models of system behavior, leading them to misinterpret outputs or rely on inappropriate heuristics when deciding whether to trust the system~\citep{kahr2024understanding,dhuliawala2023diachronic}. 
In practice, this can manifest as users treating confidence signals or fluent outputs as reliable evidence of correctness, even when these signals are difficult to calibrate across tasks and interaction histories~\citep{li2026learning,kahr2024understanding}. 
At the same time, AI systems lack the reputational histories, role expectations, and accountability structures that help people contextualize errors in human collaborators, making unexpected failures harder to interpret or recover from~\citep{lee2004trust,hoff2015trust,Glikson2020-gj,parasuraman1997humans}. 
As a result, efficient outputs can mask uncertainty, brittleness, or dataset-specific artifacts that users cannot easily anticipate or inspect.

\textbf{Extension 2: Realizing complementarity requires new forms of coordination with AI.} 
As in human-human collaboration, complementarity in human-AI collaboration requires deciding which tasks should be performed by the human, which should be performed by the AI system and how uncertainty should be communicated~\citep{rastogi2023taxonomy,Mozannar2020-tg,hemmer2025complementarity}. 
AI systems also change the coordination problem because they can shape both the work product and the process by which people coordinate around it. In production roles, AI can draft, retrieve, summarize, classify, write code, and check work, changing the distribution of cognitive labor across a workflow~\citep{Jarrahi2018-cd, Maedche2019-se,peng2023impact,ziegler2024measuring}. 
In coordination roles, AI has the potential to maintain shared representations of evidence, assumptions, model limitations, data provenance, and decision rationales, making it easier for participants to see what has been considered and what remains uncertain~\citep{mitchell2019model,pushkarna2022data,Klein2005-gr,Horvitz1999-mi}. 
These roles parallel transactive memory and shared mental models in human teams, but they require explicit design~\citep{Dafoe2021-ea,Klein2005-gr,suchman1987plans,hollan2000distributed,amershi2019guidelines}.

\begin{figure}[t]
\centering
\includegraphics[width=0.85\textwidth]{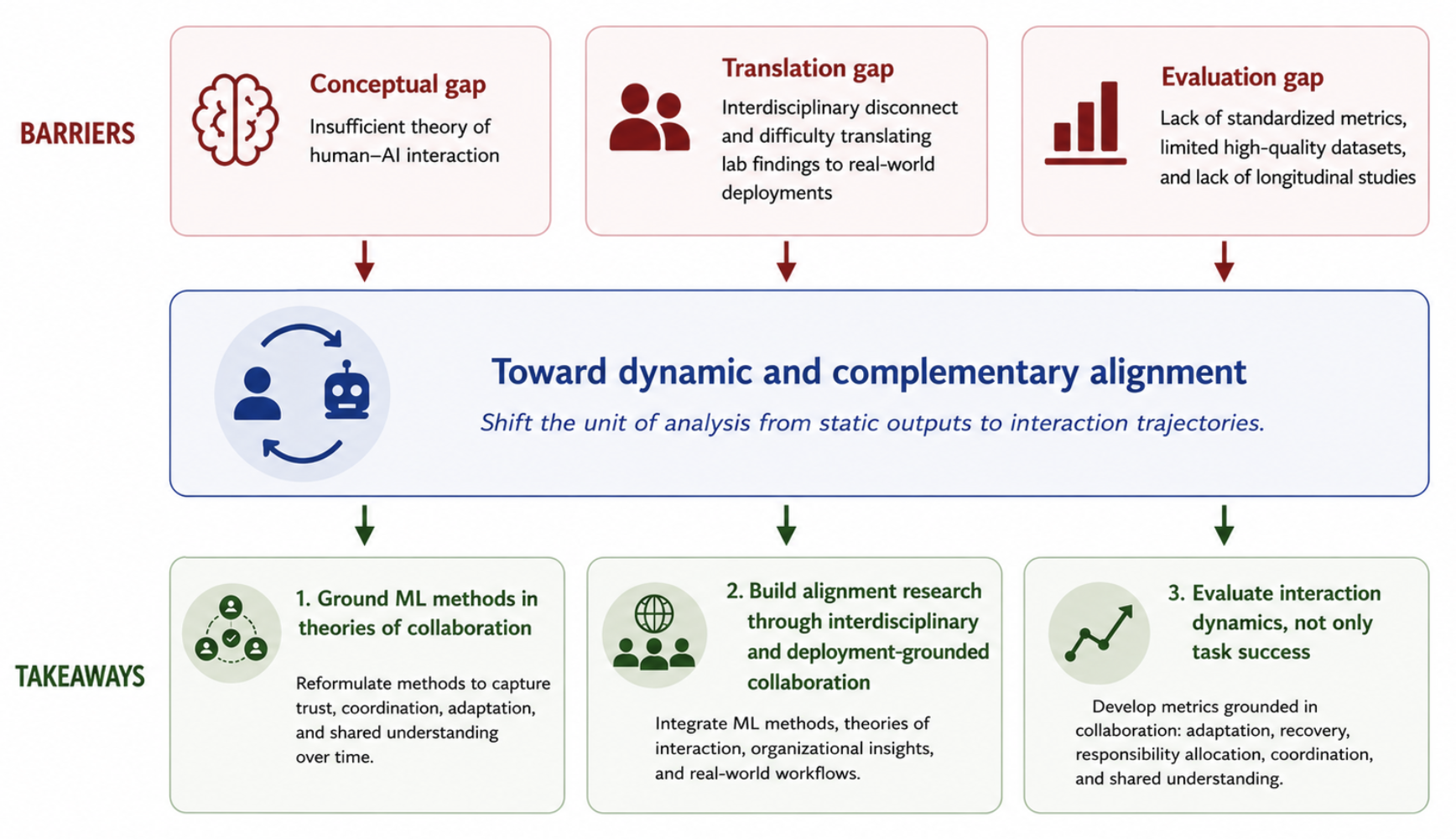}
\caption{\textbf{Bridging Barriers to Dynamic Human-AI Alignment.} Participants in the workshop noted three core barriers to move from static, output-centric views of alignment toward a trajectory-level perspective and recommended a set of takeaways.}
\label{fig:takeaways}
\end{figure}

\section{Toward Dynamic, Complementary Alignment}

To identify paths toward progress, we first examine participants' perceived barriers to advancing alignment for human-AI collaboration. We then discuss existing building blocks from the machine learning community that can help address these challenges, allowing us to synthesize a set of takeaways and future directions for the human-AI alignment community, summarized in Figure~\ref{fig:takeaways}.

\subsection{Barriers to Progress}

\textbf{Conceptual barriers.}
The most common barrier identified by participants was that there is insufficient theory of human-AI interaction. 
One contributing factor towards this might be that insights currently exist across human-computer interaction, cognitive science, and organizational psychology, which is also exemplified through the diversity of participants from computer science and social science who attended the workshop. 
Compared to more established areas such as social and organizational psychology, which have decades of theory on human teamwork and coordination, research on human-AI interaction is relatively nascent~\citep{Kozlowski2006-wz,Marks2001-zc,Fiore2016-jl}. 
Participants also noted that this challenge is compounded by the rapid evolution of AI systems themselves.

\textbf{Translation barriers.}
A second cluster of barriers reflected challenges in grounding human-AI alignment research in real-world settings. 
Participants emphasized that understanding effective human-AI collaboration often requires studying systems in deployment to surface issues that were previously not present in model-centered evaluation~\citep{holstein2019improving,sambasivan2021data}.
While there are increasingly many examples of such systems in practice, these deployments are frequently concentrated in private or industry settings, where interaction data are not publicly accessible. 
Participants also highlighted gaps in interdisciplinary collaboration between theoretical and application-grounded perspectives~\citep{grudin2009ai}. 

\textbf{Evaluation barriers.}
A third set of barriers fell under the broad umbrella of shared evaluation infrastructure. 
Participants highlighted the lack of standardized metrics, limited access to high-quality interaction datasets and interactive testbeds, and the scarcity of longitudinal studies as major obstacles to progress. 
While growing efforts have emphasized documentation and accountability practices for datasets and models~\citep{gebru2021datasheets,mitchell2019model,pushkarna2022data}, participants noted that comparable infrastructure for studying human-AI interaction remains underdeveloped.

\subsection{Existing Building Blocks in ML Literature}

The barriers identified by workshop participants do not imply that the machine learning community is starting from scratch:

 \begin{itemize}[leftmargin=15pt]

    \item \textbf{Multi-Agent Reinforcement Learning.} Multi-agent RL studies how interacting agents learn to coordinate, communicate, and divide responsibilities under shared or mixed objectives~\citep{foerster2016learning,lowe2017multiagent,Carroll2019-human}. Emerging work on collaborative interaction optimization extends this framing to language-model systems that proactively elicit intent and optimize long-term interaction success~\citep{wu2025collabllm}. These directions introduce important concepts for modeling complementary expertise and interaction dynamics, but most existing work assumes simulated agents, well-specified objectives, or bounded interaction settings, leaving many challenges of real-world human-AI collaboration unresolved.

    \item \textbf{Preference Learning.} Modern alignment methods learn from human preferences over model outputs, enabling systems to optimize for helpfulness, harmlessness, and instruction following~\citep{christiano2017deep,ouyang2022training,Stiennon2020-hf,rafailov2023direct}. More recent extensions, including multi-turn RLHF, expand these objectives from isolated responses to trajectories of interaction~\citep{lu2024multi}. However, these approaches still primarily optimize conversational preferences and assistant behavior, rather than the broader dynamics of collaboration that emerge in human-AI workflows over time.
    
    \item \textbf{Human Data Requirements.} Recent work increasingly curates datasets and benchmarks from real user interactions, including large-scale chatbot logs, software engineering tasks derived from GitHub issues, and coding-agent interaction traces~\citep{zheng2023lmsys,zhao2024wildchat,jimenez2024swebench,swechat2026,chen2025can}. These resources make evaluation more grounded in deployment data, but they still primarily capture task distributions, conversation traces, or agent capabilities, rather than directly validating whether model behavior improves longitudinal human-AI collaboration in real workflows.
    
 \end{itemize}

\subsection{Directions for the human-AI alignment community}

We posit that the central challenge of building towards dynamic, complementary alignment is not the absence of relevant methods, but rather the absence of formulations that ground them in human modeling and social science insights: 

\textbf{Addressing conceptual barriers.}
Prior work on human teamwork shows that collaboration depends on evolving trust, complementary expertise, coordination, and shared representations developed over interaction. 
These observations suggest that existing ML methods may need to be reformulated around richer models of collaborative interaction.
For example, a fruitful direction is to extend preference learning methods beyond explicit judgments over isolated outputs toward interaction-level properties such as calibrated reliance, coordination, and adaptation over time. This also raises new challenges around supervision and data collection, including whether such signals should be elicited directly from users or inferred implicitly from interaction behavior and workflow trajectories.

Similarly, existing multi-agent reinforcement learning formulations often optimize shared task reward without explicitly modeling complementary expertise or evolving shared understanding between collaborators. 
One promising direction is to incorporate ideas from transactive memory and shared mental models into these formulations by learning representations over collaborator capabilities, responsibilities, and uncertainty that evolve throughout interaction. 
At the same time, applying these methods to human-AI collaboration introduces practical challenges, as reinforcement learning methods are often highly data-intensive, while human interaction data are expensive to collect and constrained by deployment settings.

\textbf{Addressing translation barriers.}
Many of the interaction phenomena discussed earlier only emerge through sustained interaction in realistic workflows.
We argue that progress requires closer collaboration across communities that study different aspects of human-AI systems. 
Machine learning research contributes methods and evaluation frameworks, HCI and cognitive science contribute theories of interaction and human behavior, organizational research contributes insights about coordination and teamwork in practice, and industry deployments provide access to real-world workflows where these dynamics can be observed at scale. 
Creating stronger mechanisms for these perspectives to inform one another can ground future alignment research more directly in realistic collaborative settings.

\textbf{Addressing evaluation barriers.}
Recent datasets and benchmarks increasingly draw from real user interactions, including chatbot logs, software engineering workflows, and agent trajectories, representing important progress toward grounding evaluation in deployment data. 
Prior work from social and decision sciences suggests that effective collaboration is often reflected not only in whether a task is completed successfully, but in how collaborators adapt to one another over time, recover from failures, allocate responsibilities, and develop shared understanding throughout interaction. 
Future evaluation frameworks should therefore explore metrics that are more directly grounded in theories of collaboration to better understand when strong model performance translates into effective human-AI collaboration in practice, along with characterizing the effect of different training paradigms.

\section{Alternative Views}

We discuss potential counterarguments and objections to dynamic human-AI alignment:

\textbf{Prioritize alignment for autonomous AI systems.}
A first alternative view is that static alignment of AI systems is limited but indispensable. 
This view would argue that the community primarily relies on snapshot evaluations because they provide scalability and reproducibility, not because the community ignores the value of long-term interaction and complementarity~\citep{kiela2021dynabench}. 
A related perspective is that alignment should focus on dynamic settings in which AI systems may operate autonomously over long time horizons or primarily interact with other AI systems. 
Our position is that human-AI alignment does not replace static alignment or other alignment objectives around autonomous AI systems.
Rather, it complements them by focusing on an increasingly important class of real-world workflows in which humans and AI systems jointly perform work over extended periods.

\textbf{Complementarity is a workflow design problem, not an alignment problem.}
A second alternative view is that research on human-AI complementarity belongs primarily to HCI, human factors, or organizational design. 
As AI systems are deployed in agentic and interactive contexts, we argue that this boundary is increasingly difficult to maintain.
In these growing deployments, the model behavior can alter what users attend to, when they verify, how they delegate, and how work is reorganized~\citep{Horvitz1999-mi,Parasuraman2000-levels}. 
Recent studies demonstrate how human-AI teams fail to outperform the best human-only or AI-only baseline unless the collaboration is deliberately structured~\citep{Vaccaro2024-te,Bansal2021-sb}. 
Thus, achieving robust complementary collaboration may require not only redesigning the surrounding workflow, but also fundamentally rethinking how models are trained, optimized, and evaluated for interactive human-AI settings.

\textbf{Optimizing for human-AI alignment creates safety concerns.}
A final view is that optimizing human-AI interactions creates new risks. 
If an AI system is rewarded for downstream workflow outcomes, it may shape user beliefs, trust, attention, or reliance in ways that users did not explicitly endorse~\citep{parasuraman1997humans,Logg2019-appreciation,dietvorst2015algorithm}. 
For example, an AI system might learn to withhold information or induce skepticism by nudging a user's behavior toward an ulterior objective under the justification of improving long-run performance~\citep{emmons2024observation}. 
Dynamic alignment therefore introduces fundamental research challenges that should be treated as central to the broader alignment agenda rather than as reasons to avoid studying these systems altogether.
In our view, the existence of these risks strengthens, rather than weakens, the case for rigorous academic study.

\section{Conclusion}

We propose a shift from static, emulative alignment toward interactive and complementary alignment. 
We contrast standard alignment objectives with a trajectory-level formulation in which human and model behavior co-evolve, and ground this perspective in insights from an interdisciplinary workshop. 
First, we draw on social science accounts of human collaboration, where we identify key dynamics such as evolving trust and coordination, and show how these are extended and transformed in human-AI settings.
We further highlight how human-AI systems introduce new asymmetries, uncertainty, and coordination challenges. 
Finally, we distill barriers and takeaways for the human-AI alignment community, outlining implications for designing and evaluating human-AI workflows.

\bibliography{example_paper}
\bibliographystyle{plain}

\appendix
\section{Workshop Details}

TODO

\end{document}